# Computational Approach to Anaphora Resolution
# in Spanish Dialogues

**Manuel Palomar**                                    MPALOMAR@DLSI.UA.ES
*Dept. Lenguajes y Sistemas Informáticos*
*Universidad de Alicante*
*Alicante, SPAIN*

**Patricio Martínez-Barco**                           PATRICIO@DLSI.UA.ES
*Dept. Lenguajes y Sistemas Informáticos*
*Universidad de Alicante*
*Alicante, SPAIN*

## Abstract

This paper presents an algorithm for identifying noun-phrase antecedents of pronouns and adjectival anaphors in Spanish dialogues. We believe that anaphora resolution requires numerous sources of information in order to find the correct antecedent of the anaphor. These sources can be of different kinds, e.g., linguistic information, discourse/dialogue structure information, or topic information. For this reason, our algorithm uses various different kinds of information (hybrid information). The algorithm is based on linguistic constraints and preferences and uses an anaphoric accessibility space within which the algorithm finds the noun phrase. We present some experiments related to this algorithm and this space using a corpus of 204 dialogues. The algorithm is implemented in Prolog. According to this study, 95.9% of antecedents were located in the proposed space, a precision of 81.3% was obtained for pronominal anaphora resolution, and 81.5% for adjectival anaphora.

## 1. Introduction

Anaphora resolution is one of the most active areas of research in Natural Language Processing (NLP). The comprehension of anaphora is an important process in any NLP system, yet it is among the toughest problems in computational linguistics and NLP. According to Hirst (1981):

> Anaphora, in discourse, is a device for making an abbreviated reference (containing fewer bits of disambiguating information, rather than being lexically or phonetically shorter) to some entity (or entities) in the expectation that the receiver of the discourse will be able to disabbreviate the reference and, thereby, determine the identity of the entity.

The reference to an entity (e.g., a pronoun) is generally called an *anaphor*, the entity to which the anaphor refers is its *referent*, and the previous reference to the same entity is the anaphor's *antecedent*. For instance, in the statement "*John$_i$ ate an apple. He$_i$ was hungry*", the pronoun *he* is the anaphor and the noun *John* is the antecedent.

An anaphoric problem can be described as lying somewhere between the resolution and the generation of anaphora, the former term being the disabbreviating of the reference and the latter being the abbreviating form of the reference to an entity. This paper focuses exclusively on the resolution of anaphora and not on their generation. Anaphora can be classified in many different ways, depending upon the particular criteria one chooses to employ. Regarding the element that carries out the reference (the anaphor), for example, clear distinctions should be made between pronominal anaphora, adjectival anaphora, definite descriptions, one-anaphora, surface-count anaphora, verbal-phrase anaphora, and time and/or location references. This paper focuses on the resolution of pronominal and adjectival anaphora.[1]

It is widely agreed that the process of resolving anaphora in natural language texts may be supported by a variety of strategies that employ different kinds of knowledge. By different kinds of knowledge we mean the various sources of information usually employed for anaphora resolution, including morphological agreement, syntactic parallelism, semantic information, discourse structure, topical knowledge, and so on.

Natural language processing (NLP), and, specifically, anaphora resolution, uses many resources and sources of information for two reasons: (1) numerous resources are available to the scientific community; and (2) humans employ many sources of information in order to resolve different linguistic phenomena.

We present an algorithm that coordinates different forms of knowledge by distinguishing between linguistic knowledge (constraints and preferences) and dialogue-structure knowledge (anaphoric accessibility space). The algorithm identifies the noun phrase to which a third-person personal or demonstrative pronoun or adjectival anaphor[2] refers in a Spanish dialogue. We call this algorithm ARDi (anaphora resolution in dialogues). ARDi was implemented in Prolog.

In Section 2 below, we present related work on anaphora resolution in dialogues. In Section 3, we suggest an annotation scheme for capturing Spanish dialogue structure. In Section 4, an accessibility space based on this annotation scheme is defined. In Section 5, we present the algorithm ARDi. Finally, an experimental study of the algorithm is presented in Section 6.

## 2. Related work on anaphora resolution in dialogues

For anaphora resolution in dialogues, a proliferation of methods based on dialogue structure (discourse-oriented approaches) have been developed. Among these, we should like to especially acknowledge the work of Grosz (1977, 1981), in which the influence of dialogue structure in anaphora resolution is justified. Grosz's work focuses specifically on task-oriented dialogues. Other studies, such as those published by Grosz *et al.* (1983, 1995), present a centering framework as a model to explain the coherence of local discourse segments in which the speaker's focus of attention is related to referring expressions. This model has achieved successful results in anaphora resolution in monologues, but would require certain modifications to be successfully applied to dialogues. Along those lines, Byron and Stent (1998) have developed extensions of the centering method for application to dialogues. They conclude that centering in dialogues is as consistent in dialogues as it is in monologues.

---

1. We have dealt exclusively with pronominal and adjectival anaphora because they appeared most frequently in the dialogues we evaluated, but our algorithm can easily be extended to other kinds of anaphora.
2. A Spanish adjectival anaphor is a kind of English one-anaphora where the word *one* is omitted. For example, *el rojo (the red [one])*.

Nevertheless, according to Strube and Hahn (1999), the crucial point of the centering model is the candidate list. Grosz *et al.* (1995) state that this list may be ordered using different factors, but they only use information about grammatical roles. However, it is difficult to define grammatical roles in free-word-order languages like German or Spanish without using semantic information.[3]

On the other hand, work carried out by Eckert and Strube (1999) on English details a method for resolving pronominal anaphora in dialogues with a precision of 66.2% and a recall rate of 68.2%. This method is based on the distribution of dialogue acts as an alternative to the centering method.

Furthermore, Martínez-Barco *et al.* (1999) emphasize the importance of discourse-topic knowledge as a complementary method for anaphora resolution in dialogues in which such knowledge is necessary for long-distance anaphora resolution.

## 3. An annotation scheme for dialogue structure

For successful anaphora resolution in dialogues, we assume that it is essential to identify dialogue structure. Therefore, we propose an annotation scheme for Spanish dialogues that is based on work carried out by Gallardo (1996), who applies the theories put forward by Sacks *et al.* (1974) concerning (conversational) turn-taking.

We use an annotation scheme based on these theories for three main reasons. First, as it is a general approach to dialogue modeling, it is applicable to all types of dialogues, including both task-oriented and information-retrieval-oriented dialogues. Consequently, the use of such a model as a basis for developing our anaphor resolution procedure allows us to apply the procedure to any type of domain, thus offering an advantage over procedures based on discourse models specific to particular domains. Second, this annotation scheme can be easily applied to automatic processes without metalinguistic considerations. Although in our work the annotation task has been performed by hand, for dialogue-based applications in which our procedure might be embedded (e.g., in dialogue management systems), annotation tasks must be performed automatically. Finally, we wanted to base our own procedure on studies of the influence of dialogue structure on anaphora resolution that were carried out by Fox (1987), whose approach, in turn, is based on that of Sacks *et al.*

According to these theories, the basic unit of conversation is the *move*, which informs the listener about an action, request, question, etc. Moves are carried out by means of *utterances*.[4] And utterances are joined together to become *turns*.

Since our work was done using spoken dialogues that had been transcribed, turns are annotated in the texts and utterances are delimited by the use of punctuation marks or by the ends of turns. Reading a punctuation mark (., ?, !, ...) allows us to recognize the end of an utterance. These tasks do not affect the anaphora-resolution process.

---

3. A possible grammatical role definition in free-word-order languages, e.g., subject, direct object, etc., can be recovered to a large extent by inspecting NP-verb agreement, NP-clitic agreement, and case or prepositional markers. This task requires adding word-order heuristics to the parser.

4. An *utterance* in a dialogue is equivalent to a sentence in a non-dialogue, although, because of the lack of punctuation marks, utterances are recognized by means of speakers' pauses.

As a result, we propose the following annotation scheme for dialogue structure:

**Turn (T)** is identified by a change of speaker in the dialogue; each change of speaker presupposes a new turn. On this point, we make a distinction between two different kinds of turns:

- An **intervention turn (IT)** is one that adds information to the dialogue. Such turns constitute what is called *the primary system of conversation*. Speakers use their interventions to provide information that facilitates the progress of the topic of conversation. Interventions may be **initiatives ($IT_I$)** when they formulate invitations, requirements, offers, reports, etc., or **reactions ($IT_R$)** when they answer or evaluate the previous speaker´s intervention. Finally, they can also be **mixed interventions ($IT_{R/I}$)**, which is a reaction that begins as a response to the previous speaker's intervention, and ends as an introduction of new information.

- A **continuing turn (CT)** represents an empty turn, which is quite typical of a listener whose aim is the formal reinforcement and ratification of the cast of conversational roles. Such interventions lack information.

**Adjacency pair (AP)** (also called **exchange**) is a sequence of turns headed by an initiation intervention turn ($IT_I$) and ended by a reaction intervention turn ($IT_R$). This form of anaphora, in which the reference appears within an adjacency pair, appears to be very common in dialogues (Fox, 1987).

**Topic (TOPIC)**. The topic must be a lexical item that is referred to frequently. According to Rocha (1998), four features are taken into account in the selection of the best candidate for a discourse topic: frequency, even distribution, position of first token, and semantic adequacy. A highly frequent element that occurs intensively in a passage of the dialogue but does not appear for long stretches is not likely to be a good choice for discourse topic. In the same way, neither is an element whose first appearance occurs a long way from the beginning the best choice. Moreover, semantic adequacy must be considered for the candidate, and it must be assessed by the annotator.

Based on the above-mentioned structure, then, the following tags are considered necessary for dialogue structure annotation: $IT_I$, $IT_R$, **CT**, **AP**, and **TOPIC**. The AP and TOPIC tags will be used to define the anaphoric accessibility space, and the remaining tags will be used to obtain the adjacency pairs. The $IT_{R/I}$ tag, representing mixed interventions, is not included since mixed interventions can be annotated as $IT_R$ plus $IT_I$. This task is done in the annotation phase.

An example of an annotated dialogue with tags is presented in Figure 1. In the dialogue, the identifier (OP) indicates the turn of a railway company employee, and the identifier (US) indicates the client's turn.

One of the most important advantages of this annotation scheme is its compatibility with most of the dialogue-annotation schemes used in dialogue systems. Notice, for instance, that the adjacency pairs show the same structure as the *conversational game* applied to task-oriented dialogues defined in the dialogue structure by Carletta *et al.* (1997). Moreover, our

| TOPIC | | tren |
| --- | --- | --- |
| | | *(train)* |
| AP1 | $IT_I$ (OP) | información de Renfe, buenos días. |
| | | *(Renfe information, good morning.)* |
| | $IT_R$ (US) | hola, buenos días. |
| | | *(hello, good morning.)* |
| | CT (OP) | hola. |
| | | *(hello.)* |
| AP2 | $IT_I$ (US) | me podéis decir algún tren que salga mañana por la tarde para ir a Monzón? |
| | | *(could you tell me about any train that leaves tomorrow evening for Monzon?)* |
| | $IT_R$ (OP) | sí, vamos, mira hay un talgo a las tres y media de la tarde. |
| | | *(yes, let's see, there is a talgo at half-past three.)* |
| AP3 | $IT_I$ (US) | sí, tiene que ser más tarde. |
| | | *(yes, it has to be later.)* |
| | $IT_R$ (OP) | más tarde. hay un intercity a las cinco y media, un expreso a las seis. y media |
| | | *(later. there is an intercity at half-past five, an express at half-past six.)* |
| AP4 | $IT_I$ (US) | el de las seis y media ¿llega a Monzón? |
| | | *(the one at half-past six, does it go to Monzon?)* |
| AP5[a] | $IT_I$ (OP) | a ver. el de las seis y media me ha preguntado ¿verdad? |
| | | *(let me see. you've asked about the one at half-past six, right?)* |
| | $IT_R$ (US) | sí. |
| | | *(yes.)* |
| | $IT_R$ (OP) | a las nueve y veinticinco. |
| | | *(twenty-five past nine.)* |
| AP6 | $IT_I$ (US) | a las nueve y veinticinco está en Monzón? |
| | | *(at twenty-five past nine it is in Monzon?)* |
| | $IT_R$ (OP) | sí |
| | | *(yes)* |
| | CT (US) | vale, pues ya está. esto ya es suficiente. |
| | | *(ok, that's it. that's enough for now.)* |
| AP7 | $IT_I$ (US) | gracias, ¿eh? |
| | | *(thank you, eh?)* |
| | $IT_R$ (OP) | muy bien a usted. hasta luego. |
| | | *(very well, thanks to you. so long.)* |

---

*a.* This adjacency pair is included within AP4.

Figure 1: Example of an annotated dialogue from *Corpus InfoTren: Person*

scheme is also compatible with those that are based on utterance functions, such as the one defined in DAMSL by Allen and Core (1997). DAMSL indicates how utterances are related to the discourse by means of *forward- and backward-looking functions*. The interpretation of these functions builds the adjacency-pair structure. Finally, our topic structure exhibits the same features as the *transaction* structure of Carletta *et al.* or the *task level* defined by Allen and Core.

## 4. Accessibility space proposal

Based upon the above-mentioned annotation, an anaphoric accessibility space is proposed for Spanish in order to resolve anaphors in the form of personal pronouns, demonstrative pronouns, and adjectival anaphors.

### 4.1 Description

According to Fox (1987), the first mention of a referent in a sequence of contexts is performed with a full noun phrase. After that, by using an anaphor the speaker displays an understanding that sequence has not been closed down. We assert that two different sequences generate most of the anaphors to be found in dialogues: the adjacency pair and the topic scope. The former generates references to any local noun phrase, and the latter generates references to the main topic of the dialogue.

Based on this, we propose that the anaphoric accessibility space for any given anaphor may be defined as the set of noun phrases taken from:

- the adjacency pair containing the anaphor, plus

- the adjacency pair preceding the adjacency pair containing the anaphor, plus

- any adjacency pair including the adjacency pair containing the anaphor, plus

- the noun phrase representing the main topic of the dialogue.

### 4.2 An automatic topic detection proposal

Several works about automatic topic detection have been published—Reynar (1999), Youmans (1991) or Hearst (1994)—. In Martínez-Barco *et al.* (1999) an automatic topic detection algorithm as applied to anaphora resolution is presented.

This algorithm selects noun phrases (NP) occurring before an anaphor. These NPs are included in a list that is then weighted. Each time the NP appears in a new turn (frequency), its weight is increased, and each time the NP does not appear in a new turn (infrequency), its weight is decreased. According to this algorithm, the dialogue topic may be determined by its salience, i.e., by determining the NP with the heaviest weight (high frequency in a short distance) occurring before an anaphor. In order to obtain this information (weight), the algorithm uses the following two coefficients:

- $C_f$: coefficient of frequency

- $C_i$: coefficient of infrequency

$C_f$ increases the salience of a referring expression when the entity appears in the current intervention turn. $C_i$ decreases the salience of expressions that appeared in previous intervention turns but not in the current one, indicating a loss of importance. Both coefficients obviously affect the salience of expressions in reflecting their frequency and their distance from the current intervention turn where the anaphor has been found. The expression with the highest salience will be the most favored candidate antecedent on the whole list and therefore the most relevant topic for the current intervention turn.

This automatic topic detection method has the following advantage over other methods: it does not obtain a single topic, but rather a list of topic candidates ordered by salience. That is important for our anaphora resolution system because, if the highest-ranked candidate does not fulfill the relevant constraints, then the next highest candidate can be tested.

Initially, values of 10 units and 1 unit, respectively, were assigned to $C_f$ and $C_i$. These values were arrived at experimentally, but further study could lead to more precise values.

## 5. Anaphora resolution in Spanish dialogues

In this section, the anaphora resolution algorithm based on a constraint and preference approach is presented.

### 5.1 Constraints and preferences as an approach to anaphora resolution

According to Dahlbäck (1991), there are at present two basic approaches in anaphora resolution: (1) the traditional approach, which generally depends upon linguistic knowledge, and (2) the discourse-oriented approach, in which the researcher tries to model complex discourse structures and then uses these structures to resolve anaphora.

Among the traditional approaches, the work of Mitkov (1998), Baldwin (1997), and Ferrández *et al.* (1999) are all based on a combination of linguistic knowledge (lexical, morphological, syntactic, and/or semantic) for the resolution of anaphora. These approaches apply linguistic knowledge, in the way of constraints and preferences, following the work of Carbonell and Brown (1988) and Rich and LuperFoy (1988), in which such systems are proposed as a technique for combining several information sources.

These approaches are based, intuitively, on the following three steps: (1) defining an anaphoric accessibility space, (2) applying constraints, and (3) applying preferences.

A constraint and preference system must define, on the one hand, the anaphoric accessibility space. That is, it must obtain a list with all the possible candidate antecedents. On the other hand, the system must also define the text segments in which the antecedent can be found. This step has a great importance for the remaining steps in the process because a definition of the anaphoric accessibility space that is too narrow results in the exclusion of valid antecedents. Likewise, a definition of the anaphoric accessibility space that is too broad results in large candidate lists, with a corresponding increase in the likelihood of erroneous anaphora resolution. Usually, anaphora resolution systems based on linguistic knowledge (Ferrández et al., 1999) define an accessibility space using $n$ previous sentences to the anaphor, where $n$ is variable according to the kind of the anaphora.

Once the list of possible candidates is defined, several constraints are applied in order to remove incompatible antecedents. The constraint system consists of conditions that must be met, and candidates that do not fulfill these conditions will not be considered possible antecedents for the anaphor. Lexical, morphological, syntactical, and semantic information are traditionally used to define the constraints.

Finally, after removing incompatible candidates, if the remaining list contains more than one antecedent, preferences are applied in order to choose a single antecedent. In this case, unlike that of constraints, preferences are associated with likelihood lower than 100%. Candidates fulfilling a preference, then, have a greater likelihood of being the antecedent than those not fulfilling it. The preference system must be designed bearing in mind that only one candidate must remain at the end. This final candidate will be proposed as the antecedent for the anaphor. Lexical, morphological, syntactic, and semantic information are usually used in order to define the preference system.

The works of Mitkov (1998) and Ferrández *et al.* (1999) show that anaphora resolution systems based on constraints and preferences can yield successful results when applied to non-dialogue texts. However, these works lack adequate proposals for the anaphoric accessibility space. Furthermore, these approaches lack consistency in the treatment of other kinds of texts, for example, dialogues.

## 5.2 The anaphora resolution algorithm (ARDi)

In this section, the intuitive algorithm for anaphora resolution in spoken dialogue systems (ARDi) is presented. ARDi operates with syntactic information provided by the SUPP partial parser (Ferrández, Palomar, & Moreno, 1998). SUPP is based on a partial representation of slot unification grammar analysis (Ferrández et al., 1999). This partial representation gives some of the utterance constituents, such as NPs, PPs, verbal chunks, and partial information about subordinated clauses. Thus, ARDi combines two kinds of knowledge about dialogues: (1) linguistic knowledge, such as lexical, morphological, and syntactic knowledge; and (2) knowledge about the dialogue's structure itself, which is based on the annotation of adjacency pairs[5] and knowledge about the topic of the dialogue (manually annotated). Figure 2 shows the anaphora resolution procedure.

ARDi is based, intuitively, on the following three steps:

1. Obtain all possible antecedents from dialogue structure and topic as follows:

    (a) take those NPs that are included in the same adjacency pair (AP) as the anaphor, and

    (b) take those NPs that are included in the previous AP to that containing the anaphor, and

    (c) take those NPs that are included in the most recent unclosed AP containing the AP containing the anaphor, and

    (d) take the topic of the dialogue

2. Discard incompatible antecedents by applying linguistic constraints, as follows:

    (a) for pronominal anaphora:

---

5. The use of adjacency pairs as dialogue units for anaphora resolution is based on the work of Sacks *et al.* (1974), in which they suggest that one form of anaphora which appears to be very common in dialogues is reference within an adjacency pair.

```
Procedure RESOLUTION (A,L(AAS))
Let ANAPHOR = the anaphor A
Let AAS = anaphoric accessibility space from A
Let LIST = a list L(AAS) of all NPs (antecedent candidates)
           from AAS
For each NP in LIST, apply constrains of morphological
           agreement between NP and ANAPHOR to obtain LIST1
end for
For each NP in LIST1, apply constrains of syntactic conditions
           between NP and ANAPHOR to obtain LIST2
end for
For all NP in LIST2, apply linguistic and discourse structural
           preferences (in the order described in step 3 below )
           until |LIST2| = 1
end for
Return LIST2
end procedure
```

Figure 2: Anaphora resolution procedure

    i. discard those antecedents that do not agree in gender, number, and person

    ii. discard the antecedents that are non-co-referent according to the following rule:

    A pronoun P is non-co-referential with a (non-reflexive or non-reciprocal) noun phrase N if any of the following conditions[6] hold:

- P and N are in the same utterance and clause, and P and N modify the head of the same NP

- P and N are in the same utterance and clause, and P does not modify the head of any NP

(b) for adjective anaphora:

    i. discard those antecedents that do not agree in gender

    ii. discard those antecedents whose head noun is not of the lexical category "COMMON"

3. If more than one antecedent is left, filter the remaining antecedents by applying the following weighted preferences:

(a) for pronominal anaphora:

    i. antecedents that are in the same AP as the anaphor (weight = 35)

---

6. C-command and minimal governing category restrictions are proposed, as formulated in Reinhart (1983). Based on these restrictions and the non-co-reference conditions of Lappin and Leass (1994), we propose conditions for NP-pronoun non-co-reference adapted for Spanish. These conditions are applied to the syntactic information provided by the partial parser. They are of great importance because we do not use semantic information in our proposal.

    ii. antecedents that are in the previous AP to that containing the anaphor (weight = 20)

    iii. antecedents that are in the most recent unclosed AP (weight = 30)

    iv. antecedents in the topic (weight = 15)

    v. antecedents that appear with the verb of the anaphor more than once (weight = 5)

    vi. antecedents that are in the same position with reference to the verb as the anaphor (before or after)(weight = 5)

    vii. antecedents that are in the same position with reference to the utterance as the anaphor (weight = 5)

    viii. the nearest antecedent to the anaphor (used when more than one candidate obtains the highest value)

(b) for adjectival anaphora:

    i. antecedents that are in the same AP as the anaphor (weight = 35)

    ii. antecedents that are in the previous AP to that containing the anaphor (weight = 10)

    iii. antecedents that are in the most recent unclosed AP (weight = 10)

    iv. antecedents in the topic (weight = 35)

    v. antecedents that share the same kind of modifiers (e.g., prepositional phrases, adjectives, and so on) (weight = 5)

    vi. antecedents with exactly the same modifiers (e.g., the same adjective 'red') (weight = 5)

    vii. antecedents that agree in number (weight = 5)

    viii. the nearest antecedent to the anaphor (used when more than one candidate obtains the highest value)

These preferences were developed as a result of the empirical study explained in the following section.

## 6. Experimental work

In this section, an experimental study of the algorithm is presented, including a deep description of the experiments, corpora and tools used, as well as a study about the importance of the anaphoric accessibility space.

### 6.1 Corpora, tools, and description of experiments

In order to evaluate the anaphora resolution algorithm proposed in this paper, the general process outlined in Figure 3 was followed.

Data for the evaluation were taken from the *Corpus InfoTren: Person*, a corpus of 204 transcribed spoken Spanish dialogues provided by the Basurde Project (Basurde Project, 1998). These dialogues are conversations between a railway company employee and a client. The transcriptor used in the Basurde Project provides turn and speaker markup. Out of 204 dialogues, 40 were selected for the training (training corpus) and the remaining 164 were reserved for the final evaluation (test corpus). These 204 dialogues contain 345 pronominal anaphors and 257 adjectival anaphors.

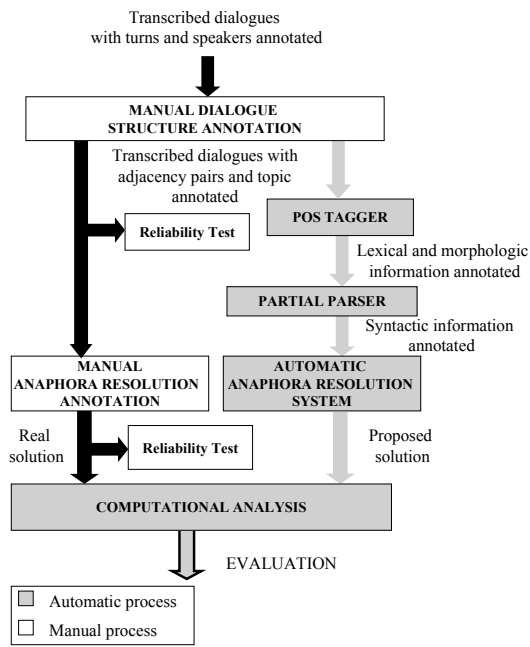

Figure 3: Full evaluation process

We had two aims in using the training corpus: (1) to estimate the importance of the structural anaphoric accessibility space, and (2) to define an adequate set of constraints and preferences (experiments 0, 1, 2, and 3). The test corpus was reserved to obtain the final evaluation.

In addition, the entire corpus was manually annotated with two different goals: (1) to identify further discourse structural properties such as adjacency pairs and topics, and (2) to identify anaphors and antecedents. Although we annotated the corpus manually, there are at present some automatic systems for performing adjacency pair tagging (the Basurde Project (Basurde Project, 1998), for example), as well as for automatic topic tagging (Reynar, 1999) or automatic topic extraction (see the method for anaphora resolution described in Section 4).

The annotation of conversational structure was carried out as described in the next paragraph. An important aspect of dialogue structure annotation is the training phase, which assures reliability among annotators.

The annotation phase was accomplished as follows: (1) two annotators were selected, (2) an agreement[7] was reached between the two annotators with regard to the annotation scheme using a training corpus, (3) the annotation was then carried out by both annotators in parallel over the test corpus, and (4) a reliability study was carried out on the annotation (Carletta *et al.* 1997). The reliability study used the *kappa* statistic that measures the affin-ity between the annotations of the two annotators by making judgments about categories. For computing the *kappa* (*k*) statistic, see Siegel and Castellan (1988).[8]

---

7. This agreement is about what every tag means to every annotator when it is applied to the corpus.

Because turns are marked during the transcription phase, the annotator merely classifies turns according to the turn types described in Section 3 and then relates each initiative intervention $IT_I$ to its reaction intervention $IT_R$, thereby defining adjacency pairs. Since this task simply requires classification, it is easily measured using the *kappa* statistic.

Concurrently, topics were identified. This task was also simple, since the corpus used for these experiments is organized into short dialogues and each dialogue has only one main topic or theme, and since these are introduced clearly by means of the client's intervention at the beginning of each dialogue. As a result, we detected no discrepancies between annotators with regard to the topic identification. Therefore, there was no need to measure this task using the *kappa* statistic.

According to Carletta *et al.*, a *k* measurement between 0.68 and 0.80 allows us to make positive conclusions, and if *k* is greater than 0.80, we have total reliability between the results of the annotators.

In those cases where a discrepancy was found between the annotators, the following criterion was applied: each dialogue was assigned a main annotator whose annotation was considered definitive in the event that there were discrepancies between the two accounts. In order to guarantee balance, each annotator was the main annotator for exactly 50% of the dialogues.

Once both annotators had finished the annotation, the reliability study was carried out, with a resultant *kappa* measurement of $k = 0.91$. We therefore consider the annotation obtained for the evaluation to be totally reliable.

Since the annotated texts would be processed by an anaphora resolution system, we developed an SGML tagging format.

Generally, this SGML markup will have the following form:

```
<ELEMENT-NAME ATTRIBUTE-NAME="VALUE" ...> text-string </ELEMENT-NAME>
```

Thus, the following notations are used in each case:

- Topic:

  ```
  <TOPIC> Topic-entity </TOPIC>
  ```

- Adjacency pairs:

  ```
  <AP ID="number"> Adjacency-pair </AP>
  ```

  where ID is an identification number used to arrange the adjacency pairs in sequential order

- Intervention turns:

  ```
  <IT TYPE="R|I" SPEAKER="speaker"> Intervention-turn </IT>
  ```

---

8. Additional information about this annotation process can be found in our detailed reports (Martínez-Barco, 2001; Martínez-Barco & Palomar, 2000).

where TYPE may be "R" or "I" (Reaction or Initiative) and SPEAKER is the indicator for the speaker whose turn it is

- Continuing turns:

```
<CT SPEAKER="speaker"> Continuing-turn </CT>
```

The format is exemplified in Figure 4.

---

| | |
|---|---|
| <TOPIC> | tren |
| | *(train)* |
| </TOPIC> | |
| | ... |
| <AP ID="4"> | |
| <IT TYPE="I" SPEAKER="CL"> | el de las seis y media ¿llega a Monzón? |
| | *(the one at half-past six, does it go to Monzon?)* |
| </IT> | |
| <AP ID="5"> | |
| <IT TYPE="I" SPEAKER="OP"> | a ver. el de las seis y media me ha preguntado ¿verdad? |
| | *(let me see. you've asked about the one at half-past six, right?)* |
| </IT> | |
| <IT TYPE="R" SPEAKER="CL"> | si |
| | *(yes)* |
| </IT> | |
| </AP> | |
| <IT TYPE="R" SPEAKER="OP"> | a las nueve y veinticinco. |
| | *(twenty-five past nine.)* |
| </IT> | |
| </AP> | |
| | ... |

---

Figure 4: Example of SGML annotation

As explained above, in addition to this structural annotation, the corpus was also anaphorically annotated by marking up the anaphoric relations between all pronominal and adjectival anaphors and their correct antecedents. In order to guarantee the results, this annotation was performed by two different annotators in parallel and a reliability study of the subsequent annotation was then carried out. Once again, the annotation was treated as a classification task, consisting of selecting the appropriate elements in the candidate list (we estimated an average of 6.5 possible antecedents per anaphor after applying constraints). The reliability study of the manual anaphoric annotation resulted in a *kappa* measurement of $k = 0.87$.

In addition, the corpus was tagged using the POS tagger technique described by Pla and Prieto (1998). From this, we obtained morphological and lexical information. The corpus was then parsed using the SUPP partial parser proposed by Martínez-Barco *et al.* (1998) in order to obtain syntactic information. Finally, the proposed anaphora resolution algorithm was applied.

| | Same AP[a] | Previous AP[b] | Included AP[c] | TOPIC[d] | Elsewhere[e] |
|---|---|---|---|---|---|
| Pronominal | 60.6% | 24.6% | 8.2% | 4.9% | 1.7% |
| Adjectival | 44.7% | 28.9% | 5.2% | 13.4% | 7.8% |
| Total Results | Anaphoric accessibility space proposed: 95.9% (pronominal: 98.3%, adjectival: 92.2%) | | | | 4.1% |

a. The antecedent is found in the same adjacency pair as the anaphor
b. The antecedent is found in the previous adjacency pair to the one containing the anaphor
c. The antecedent is found in the adjacency pair containing the adjacency pair including the anaphor
d. The antecedent is found in the topic of the dialogue
e. The antecedent is found elsewhere

Table 1: Structural anaphoric accessibility space results

Several studies were then carried out in order to identify the importance of defining an adequate anaphoric accessibility space and of defining a constraint and preference system based on this space. In these studies, we compared the output of the anaphora resolution system with the manual annotation and generated several statistical results.

## 6.2 Importance of the anaphoric accessibility space

In order to show the importance of defining an adequate anaphoric accessibility space, a study of the location of the antecedent of each pronominal and adjectival anaphora was done using the training corpus. The results are given in Table 1.[9]

As can be seen in the table, 95.9% of the antecedents were located in the proposed structural anaphoric accessibility space. It is estimated that the remaining antecedents (4.1%) are located in the subtopics of the dialogues.[10] In order to incorporate these remaining antecedents into the anaphoric accessibility space, one might employ a strategy that uses the **full space** (i.e., all the noun phrases from the beginning of the dialogue to the anaphor might be used). However, as shown in Table 2, our proposal for the anaphoric accessibility space (hereafter referred to as **structural**), reduces the average number of candidates per anaphor (before applying constraints) to 10.74 from the 34.14 that would be obtained if the **full space** approach were adopted. In others words, using the **full space** approach would increase the number of possible candidates by a factor of three, thereby greatly increasing both the required computational effort and the possibility of selecting incorrect antecedents. Notice, too, that these experiments were performed over a collection of short dialogues (around 332 words per dialogue). These problems will be even more acute in longer dialogues.

Other researchers have proposed using a window with a fixed number of sentences to define the anaphoric accessibility space. This type of approach might be called a **window of sentences** approach. For example, Ferrández *et al.* (1999) propose using the three previous sentences to define the accessibility space for pronouns and the four previous sentences for adjectival anaphora in Spanish. For English, Kameyama (1997) proposes the same space for the pronominal. However, there is no structural justification for these definitions. Ferrández

9. Notice that the study of the anaphoric accessibility space was not carried out using output from the automatic anaphora resolution system but rather from the manual annotations (i.e., the correct solutions).
10. By subtopic, we mean an NP that is not the main topic of the dialogue but contributes to defining it.

| Anaphoric accessibility space | Structural | Full space | Window of utterances |
|---|---|---|---|
| Total candidates | 1,063 | 3,380 | 1,292 |
| Candidates per anaphor | 10.74 | 34.14 | 13.05 |
| Proportion | 100% | 318% | 122% |

Table 2: Candidates to be processed for each anaphoric accessibility space

*et al.* and Kameyama performed several empirical studies to show the optimal space for each experiment. Table 3 below shows the results of a study which we performed using the *Corpus Infotren: Person*, the goal of which was to define an anaphoric accessibility space based on a **window of sentences** that can then be adapted to dialogues by means of a **window of utterances**. As the table shows, 11 utterances for pronominal anaphora and 10 utterances for adjectival anaphora are needed in order to cover the same number of antecedents as was covered using the structural anaphoric accessibility space (which was defined based on adjacency pairs and the topic). Since the anaphoric space using a window of utterances is not based on any principle, but rather on empirical studies, it may vary from one text to another and therefore is inadequate. Moreover, the structural anaphoric accessibility space can cover only those cases that refer to NPs introduced at the outset of the dialogue (topics), not those with a window of sentences/utterances approach.

In conclusion, it would appear that the **structural** anaphoric accessibility space is to be preferred, at least for anaphora resolution in dialogues.

| **Window of utterance**: *"From Anaphor's utterance to"*: | % pronominal anaphora antecedents | % adjectival anaphora antecedents |
|---|---|---|
| Anaphor's utterance | 37.7 | 18.4 |
| -1 | 54.1 | 44.7 |
| -2 | 70.5 | 52.6 |
| -3 | 77.0 | 55.3 |
| -4 | 80.3 | 57.9 |
| -5 | 83.6 | 71.0 |
| -6 | 88.5 | 73.7 |
| -7 | 91.8 | 76.3 |
| -8 | 91.8 | 81.6 |
| -9 | 95.1 | 81.6 |
| -10 | 96.7 | **92.1** |
| -11 | **98.4** | 94.7 |
| -12 | 98.4 | 97.4 |
| -13 | 98.4 | 97.4 |
| -14 | 100.0 | 100.0 |

Table 3: Empirical study of anaphoric accessibility space based on a window of utterances

| Experiment | Used preferences | | | | | | | | | | | | | | Precision |
|---|---|---|---|---|---|---|---|---|---|---|---|---|---|---|---|
| | Pronominal anaphora | | | | | | | | | | | | | | |
| No. | 1 | 2 | $2_a$ | $2_b$ | 3 | 4 | 5 | 6 | 7 | 8 | 9 | 10 | 11 | 12 | % |
| 0 | • | • | | | • | • | • | • | • | • | • | • | • | • | 59.0 |
| 1 | • | • | • | • | | | | | | | | | | • | 62.3 |
| 2 | • | • | • | • | | | | | • | • | • | | | • | 73.8 |
| $3^a$ | • | • | • | • | | | | | | • | | | | • | 81.3 |
| Experiment | Adjectival anaphora | | | | | | | | | | | | | | |
| No. | 1 | 2 | $2_a$ | $2_b$ | 3 | 4 | 5 | 6 | 7 | 8 | | | | | % |
| 0 | • | • | | | • | • | • | • | • | | | | | | 23.7 |
| 1 | • | • | • | • | | | | | • | | | | | | 65.8 |
| 2 | • | • | • | • | • | • | • | | • | | | | | | 78.9 |
| $3^b$ | • | • | • | • | • | • | • | | • | | | | | | 81.5 |

a. Preference weighted management
b. Preference weighted management

Table 4: Experiment summary

## 6.3 Constraint and preference set

As for the importance of defining an adequate constraint and preference set based on dialogue structure using the accessibility space defined in Section 4, we begin by adopting the constraint and preference set developed by Ferrández *et al.* (1999) and described below in Section 6.4. This constraint and preference set has been shown to be adequate for pronominal and adjectival anaphora in non-dialogue discourse. To this set, information about dialogue structure will be applied in order to take advantage of its influence. Not only is dialogue structure used to define the anaphoric accessibility space, but it is used to define preferences as well.

For this study, several experiments were carried out using the training corpus. These experiments involved changes in the constraint and preference set in order to define the configuration[11] that would have optimum precision. Results are summarized in Table 4.

### 6.3.1 PREFERENCE MANAGEMENT

There are two different approaches to managing the preference set, ordered management and weighted management. Ordered management is based on discarding those antecedents that do not fulfill a preference if there is any candidate that fulfills it. Weighted management is based on assigning a weight to each preference and then selecting the candidate with the maximum value.

For these experiments, the system was trained so as to obtain the best set of preferences. Subsequently, we applied both of the approaches to preference management to obtain the best result with the training corpus. Once we obtained the best set of preferences and its best management, we evaluated the system with the test corpus in order to obtain independent results.

---

11. The term *configuration* is used here to define the set of constraints and preferences that makes up the system used for a concrete instance of the experimental process.

So, for experiments 0, 1, and 2, ordered management was applied to obtain the best set of preferences. Then, in experiment 3, we applied weighted management to improve the results in the training corpus.

## 6.4 Experiment 0 (baseline): Linguistic information only

We began with the constraint and preference set used by Ferrández *et al.* (1999). This algorithm is based on linguistic information only, and its results have been successfully tested over a non-dialogue corpus with a resulting precision of 82% for pronominal anaphora resolution (we have no information about the precision for adjectival anaphora).

The initial configuration included the following constraint and preference set and definition of anaphoric accessibility space.

### 6.4.1 ANAPHORIC ACCESSIBILITY SPACE

For pronominal anaphora resolution, the anaphoric accessibility space consisted of the three previous turns to the anaphor. For adjectival anaphora, the space consisted of the previous four turns.

### 6.4.2 CONSTRAINTS

- In the case of pronominal anaphora, the constraints included:

  1. morphological agreement: discard the antecedents which are incompatible morphologically (gender, number, and person)
  2. syntactic context: discard the antecedents which are non-co-referent according to Lappin and Leass (1994)

- In the case of adjectival anaphora, the constraints included:

  1. morphological agreement: discard the antecedents which are incompatible morphologically (gender)
  2. Proper-noun-phrase exclusion: exclude noun phrases having a proper noun[12] as head

### 6.4.3 PREFERENCES

- In the case of pronominal anaphora, the preferences are for:

  1. candidates in the same turn as that of the anaphor
  2. candidates in the previous turn
  3. candidate proper nouns or indefinite NPs

---

12. Proper nouns are not usually modified by adjectives. Thus, noun phrases having a proper noun as head are not likely candidates as antecedents of adjectival anaphors.

4. (for personal pronouns) candidate proper nouns

5. candidates that have been repeated more than once (repeated forms and repeated mentions)

6. candidates that have appeared more than once in construction with the verb in construction with the anaphor

7. candidates in the same position as the anaphor with reference to the verb (before or after)

8. candidates in the same position with reference to the utterance as the anaphor

9. candidates not in circumstantial adjuncts

10. candidates most repeated in the text

11. candidates most often appearing in construction with the verb in construction with the anaphor

12. the closest candidate to the anaphor

- In the case of adjectival anaphora, the preferences are for:

1. candidates in the same turn as that of the anaphor

2. candidates in the previous turn

3. candidates sharing the same kind of modifier as the anaphor (e.g., a prepositional phrase)

4. candidates sharing the same modifier as the anaphor (e.g., the same adjective: 'roja' red)

5. candidates agreeing in number

6. candidates most often repeated in the text

7. candidates appearing most often in construction with the verb in construction with the anaphor

8. the closest candidate to the anaphor

### 6.4.4 DISCUSSION

Given this first configuration, an evaluation was carried out which resulted in a precision of 59.0% for pronominal anaphora resolution and 23.7% for adjectival anaphora resolution. Needless to say, these results are very low for pronominal anaphora and extremely poor for adjectival anaphora. In evaluating the errors, we concluded that the defined anaphoric accessibility space was too constrained and too arbitrarily defined. It simply disregarded the relationship between anaphora and dialogue structure. Consequently, we proposed the following changes for the second experiment.

## 6.5 Experiment 1: Dialogue structure information only

For this experiment, the definition of the anaphoric accessibility space was changed to employ information provided by the dialogue structure, as suggested by Martínez-Barco (1999). In addition, the preferences affected by this revised definition of anaphoric accessibility space were modified as well.

### 6.5.1 ANAPHORIC ACCESSIBILITY SPACE

The adjacency pair and the topic of the dialogue were used in order to define the anaphoric accessibility space. Concretely, we defined an anaphoric accessibility space by means of the adjacency pair of the anaphor, the previous adjacency pair to the adjacency pair of the anaphor, adjacency pairs containing the adjacency pair of the anaphor, and, finally, the main topic of the dialogue (for pronominal as well as adjectival anaphora).

### 6.5.2 PREFERENCES

For Experiment 1, we removed the pronominal anaphora preferences described above in items 3 through 11 and the adjectival anaphora preferences described in items 3 through 7. Also, preferences 1 and 2 above were replaced by the following four new preferences. Thus, both for pronominal and adjectival anaphora, the preferences are for:

- 1. candidates in the same adjacency pair as that of the anaphor

- 2. candidates in the previous adjacency pair to the anaphor

- $2_a$. candidates in any adjacency pair containing the adjacency pair of the anaphor

- $2_b$. candidates that are in the topic

This change was made in order to test the system's performance when linguistic information is removed and only dialogue structure information is used (preferences 1 through $2_b$). In order to guarantee a single final solution, only linguistic preference item 12, for pronominal anaphora, and item 8, for adjectival anaphora (the closest candidate), remain.

### 6.5.3 DISCUSSION

After including information about dialogue structure and removing the linguistic preferences, precision rates rose to 62.3% for pronominal anaphora resolution and 65.8% for adjectival anaphora resolution. A considerable increase is gained in the resolution of adjectival anaphora by simply changing the definition of the accessibility space. That is due to the fact that adjectival anaphora need a larger space than that used in Experiment 0.

But these results are still low and demonstrate that dialogue structure information alone is not sufficient. Thus, a third experiment was carried out using both dialogue structure and linguistic information. Several variations in the preference system were investigated independently.

## 6.6 Experiment 2: Linguistic information plus dialogue structure information (with ordered management of preferences)

Following, the preferences used in this experiment and their justification are shown.

### 6.6.1 PREFERENCES

In this experiment, we first used a preference set that included all the linguistic and dialogue structure preferences described above. Then, various alternatives were used in order to obtain an optimal configuration. As a result, the following preferences were arrived at for the final configuration:

- For pronominal anaphora, the preferences are for:

    - dialogue structure preferences: 1 through $2_b$
    - linguistic preferences: 6, 7, 8, and 12

- For adjectival anaphora, the preferences are for:

    - dialogue structure preferences: 1 through $2_b$
    - linguistic preferences: 3, 4, 5, and 8

This final set of constraints and preferences is the one that is presented in Section 5.2.

### 6.6.2 DISCUSSION AND JUSTIFICATION

It should be noted that information about repeated candidates—for example, the pronominal anaphora preferences 5, 10, and 11, or the adjectival anaphora preferences 6 and 7—is usually inserted into the preference system in order to achieve knowledge about the main entities of the dialogue. However, in this experiment, information about the main topic of the dialogue has been included and so information about repeated candidates is unnecessary. Those preferences were therefore removed, improving the results.

Furthermore, we found that pronominal anaphora preferences 3 and 4 for proper nouns caused errors. This is because in the domain of the experiment there is an exaggerated use of place names where these preferences incorrectly apply. By removing them, better results were obtained.

Finally, since the usefulness of preference 9 (candidates that are not in circumstantial adjunct) has never been justified properly, it too was omitted. After removal, the precision for pronominal anaphora stayed the same.

Thus, having considered all possible applications for ordered preference management, and given that this final set of preferences represented the minimum set of preferences, we considered it to be the optimal set. We then applied this optimal set of preferences to the training corpus, obtaining a precision of 73.8% for pronominal anaphora resolution and 78.9% for adjectival anaphora.

## 6.7 Experiment 3: Linguistic information plus dialogue structure information (with weighted management of preferences)

Following, the preferences used in this experiment and their justification are shown.

### 6.7.1 PREFERENCES

In this final experiment, the preference set obtained in the previous experiment was used (including the minimum set of preferences that we considered to be the optimal configuration). Then, several alternatives were used in order to obtain an optimal preference weight assignment. Tables 5 and 6 show the preference weight assignments for pronominal and adjectival anaphora, respectively.

This final set of constraints and preferences was presented in Section 5.2.

| Pref. No. | Description | Weight |
|---|---|---|
| 1 | Antecedents that are in the same AP as the anaphor | 35 |
| 2 | Antecedents that are in the previous AP to that containing the anaphor | 20 |
| $2_a$ | Antecedents that are in the most recent unclosed AP | 30 |
| $2_b$ | Antecedents in the topic | 15 |
| 6 | Antecedents that appear with the verb of the anaphor more than once | 5 |
| 7 | Antecedents that are in the same position with reference to the verb as the anaphor (before or after) | 5 |
| 8 | Antecedents that are in the same position with reference to the utterance as the anaphor | 5 |
| 12 | The nearest antecedent to the anaphor | YES |

Table 5: Preference weight assignment for pronominal anaphora

| Pref. No. | Description | Weight |
|---|---|---|
| 1 | Antecedents that are in the same AP as the anaphor | 35 |
| 2 | Antecedents that are in the previous AP to that containing the anaphor | 10 |
| $2_a$ | Antecedents that are in the most recent unclosed AP | 10 |
| $2_b$ | Antecedents in the topic | 35 |
| 3 | Antecedents that share the same kind of modifiers | 5 |
| 4 | Antecedents with exactly the same modifiers | 5 |
| 5 | Antecedents that agree in number | 5 |
| 8 | The nearest antecedent to the anaphor | YES |

Table 6: Preference weight assignment for adjectival anaphora

### 6.7.2 DISCUSSION AND JUSTIFICATION

In order to obtain the optimal preference weight assignment, we performed several tests with the training corpus. Thus, after looking at all the possibilities and given that this was the set of preferences having the best results, we considered it to be the optimal configuration. With this configuration, we obtained a precision of 80.3% for pronominal anaphora resolution and 92.1% for adjectival anaphora resolution.

## 6.8 Final evaluation (test corpus)

Using the final preference set defined in Experiment 3 and the proposed constraint set, a blind evaluation was carried out over the entire test corpus. This evaluation was performed independent of the training process so as to guarantee that the training would have no influence over the final percentages.

As a result, we obtained a precision of 81.3% for pronominal anaphora resolution and 81.5% for adjectival anaphora resolution.

## 7. Conclusion

In this paper we have presented an algorithm for identifying the noun phrase antecedents of pronouns and adjectival anaphors in Spanish dialogues. This algorithm exploits different kinds of information: linguistic knowledge, discourse/dialogue structure information, and discourse topic knowledge. It is based on a set of constraints and preferences which depend on all available knowledge in order to resolve anaphora.

In addition, a definition of the anaphoric accessibility space based on discourse/dialogue structure information was presented. We have shown the importance of this accessibility space in anaphora resolution, in contrast to algorithms that do not rely on any such space. Results show that 95.9% of the antecedents were located in the proposed space.

Finally, we described a set of experiments concerning this algorithm and accessibility space using a corpus of 204 dialogues. The algorithm was implemented using Prolog. In our final experiment, a precision of 81.3% was achieved for pronominal anaphora resolution and a precision of 81.5% was achieved for adjectival anaphora resolution.

As a tool for resolving pronominal and adjectival anaphora for Spanish dialogues, this system can be used in support of various NLP tasks, including machine translation, information extraction, retrieval information, or question-answering.

Currently, the authors are working on incorporating semantic information into the algorithm.

## Acknowledgments

The authors wish to thank Natividad Prieto, Ferran Pla, and Antonio Molina for having contributed their tagger; Lidia Moreno for her helpful revisions of the ideas presented in this paper; and Rafael Muñoz, Maximiliano Saiz-Noeda, Antonio Ferrández, and Jesús Peral for their collaboration in performing the experiments. We are also grateful to several anonymous reviewers of the *Journal of Artificial Intelligence Research* for helpful comments on earlier drafts of the paper.

This research has been supported by the Comisión Interministerial de Ciencia y Tecnología (CICYT) of the Spanish government, under project numbers TIC97-0671-C02-01/02 and HB1998-0068.